\documentclass{article}

\usepackage[utf8]{inputenc}
\usepackage{amsmath}
\usepackage{geometry}
\usepackage{natbib}
\usepackage{graphicx}
\usepackage{caption}
\usepackage{subcaption}
\usepackage{setspace}
\usepackage{tabularx}
\usepackage{framed}
\usepackage{epigraph}
\usepackage{booktabs}
\usepackage{color}
\usepackage{hyperref}
\hypersetup{
    colorlinks,
    citecolor=black,
    filecolor=black,
    linkcolor=black,
    urlcolor=black
}
\usepackage{framed}

\newif\ifshowcomments
\showcommentstrue
\ifshowcomments
\newcommand\zxie[1]{\textcolor{red}{[zxie: #1]}}
\else
\newcommand\zxie[1]{}
\fi

\newcommand*{\thead}[1]{\bfseries #1}
\newcommand{\unk}{$\langle\mathrm{unk}\rangle$}
\newcommand{\sos}{$\langle\mathrm{sos}\rangle$}
\newcommand{\mathrmsos}{\langle\mathrm{sos}\rangle}
\newcommand{\eos}{$\langle\mathrm{eos}\rangle$ }

\newcommand\titletext{Neural Text Generation: A Practical Guide}
\title{\titletext}

\begin{document}

\begin{center}
  {\LARGE \titletext}\\
  \vspace{1em}
  Ziang Xie\\
  Department of Computer Science\\
  Stanford University\\
  \texttt{zxie@cs.stanford.edu}
\end{center}

\setstretch{1.15}
\vspace{1em}

\begin{center}
\today\\
Most recent revision at:\\
\url{http://cs.stanford.edu/~zxie/textgen.pdf}
\end{center}

\vspace{1em}

\begin{abstract}
Deep learning methods have recently achieved great empirical success on
machine translation, dialogue response generation, summarization, and other
text generation tasks.
At a high level, the technique has been to train end-to-end neural network
models consisting of an encoder model to produce a hidden representation of
the source text, followed by a decoder model to generate the target.
While such models have significantly fewer pieces than
earlier systems, significant tuning is still required to achieve good performance.
For text generation models in particular, the decoder can behave in undesired
ways, such as by generating truncated or repetitive outputs, outputting
bland and generic responses, or in some cases producing ungrammatical gibberish.
This paper is intended as a practical guide for resolving such undesired behavior in
text generation models, with the aim of helping enable real-world
applications.

\vfill
\end{abstract}

\pagebreak

\setlength{\epigraphwidth}{25em}
\renewcommand{\epigraphflush}{center}
\epigraph{[Language makes] infinite use of finite means.}{\emph{On Language}\\Wilhelm von Humboldt}
\epigraph{But Noodynaady's actual ingrate tootle is of come into the garner mauve and thy nice are stores of morning and buy me a bunch of iodines.}{\emph{Finnegans Wake}\\James Joyce}

\pagebreak

\tableofcontents
\pagebreak

\section{Introduction}

\begin{figure}[h]
  \centering
  \includegraphics{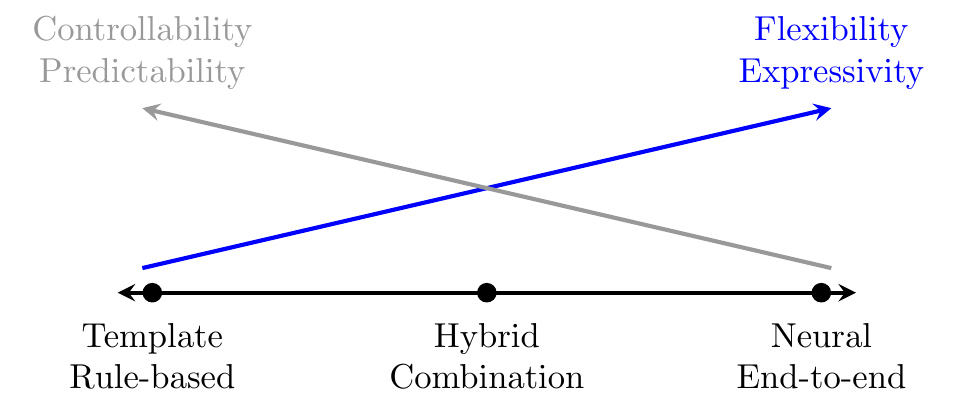}
  \caption{Figure illustrating the tradeoffs between using rule-based vs.
    neural text generation systems.}
  \label{fig:spectrum}
\end{figure}

Neural networks have recently attained state-of-the-art results on many tasks in machine learning,
including natural language processing tasks such as sentiment understanding
and machine translation. Within NLP, a number of core tasks involve
generating text, conditioned on some input information.
Prior to the last few years, the predominant techniques for text generation were either based
on template or rule-based systems, or well-understood probabilistic models
such as $n$-gram or
log-linear models~\citep{chen1996empirical,koehn2003statistical}.
These rule-based and statistical models, however,
despite being fairly interpretable and well-behaved,
require infeasible amounts of hand-engineering to scale---in the case of rule
or template-based models---or tend to saturate in their
performance with increasing training data~\citep{jozefowicz2016exploring}.
On the other hand,
neural network models for text,
despite their sweeping empirical success, are
poorly understood and sometimes poorly behaved as well.
Figure~\ref{fig:spectrum} illustrates the trade-offs between
these two types of systems.

To help with the adoption of more usage of neural text generation systems, we detail
some practical suggestions for developing NTG systems.
We include a brief overview of both the training and decoding procedures, as
well as some suggestions for training NTG models. The primary focus, however,
is advice for diagnosing and resolving pathological behavior during
decoding.
As it can take a long time to retrain models, it is comparatively cheap to
tune the decoding procedure; hence it’s worth understanding how to do this
quickly before deciding whether or not to retrain.
\begin{figure}[h]
  \centering
  \includegraphics[scale=1.1]{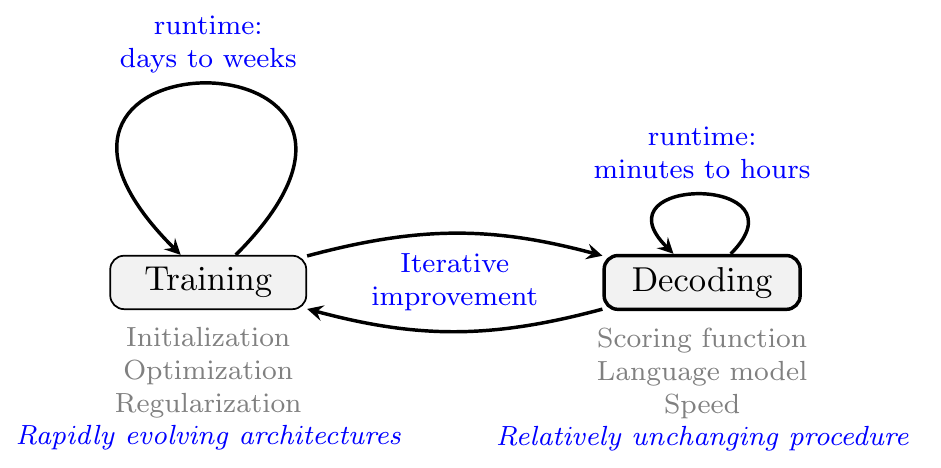}
  \caption{Development cycle for NTG systems.}
  \label{fig:train_decode_loops}
\end{figure}
Figure~\ref{fig:train_decode_loops}
illustrates the feedback loops when improving different components of the
model training and decoding procedures.

Despite a growing body of research, information on best practices tends to
be scattered and often depends on specific model architectures.
While some starting hyperparameters are suggested, the advice in this guide
is intended to be as architecture-agnostic as possible,
and error analysis is emphasized instead.
It may be helpful to first read the background section,
but the remaining sections can be read independently.

\subsection{Focus of this guide}

This guide focuses on advice for training and decoding of neural encoder-decoder models
(with an attention mechanism) for text generation tasks. Roughly speaking,
the source and target are assumed to be on the order of dozens of tokens.
The primary focus of the guide is on the decoding procedure. Besides suggestions
for improving the model training and decoding algorithms, we also touch briefly
on preprocessing (Section~\ref{sec:preproc}) and deployment (Section~\ref{ssec:deploy}).

\subsubsection{Limitations: What will \emph{not} be covered}

Before continuing, we describe what this guide will \emph{not} cover, as
well as some of the current limitations of neural text generation models.
This guide does not consider:
\begin{itemize}
  \item Natural language understanding and semantics.
    While impressive work has been done in learning word embeddings \citep{mikolov2013distributed,pennington2014glove}, the goal of learning ``thought vectors'' for sentences has remained more elusive~\citep{kiros2015skip}. As previously mentioned, we also do not consider sequence labeling or classification tasks.
  \item How to capture long-term dependencies (beyond a brief discussion of attention) or maintain global coherence. This remains a challenge due to the curse of dimensionality as well as neural networks failing to learn more abstract concepts from the predominant next-step prediction training objective.
  \item How to interface models with a knowledge base, or other structured data that cannot be supplied in a short piece of text. Some recent work has used pointer mechanisms towards this end~\citep{vinyals2015pointer}.
  \item Consequently, while we focus on natural language, to be precise, this guide does not
			cover \emph{natural language generation} (NLG), which
			entails generating documents or longer descriptions from structured data.
			The primary focus is on tasks where the target is a single sentence---hence the term ``text generation"
			as opposed to ``language generation''.
\end{itemize}

Although the field is evolving quickly, there are still many tasks where
older rule or template-based systems are the only reasonable option.
Consider, for example, the seminal work on
ELIZA~\citep{weizenbaum1966eliza}---a computer program intended to emulate a psychotherapist---that
was based on pattern matching and rules for generating responses.
In general, neural-based systems are unable perform the dialogue state management
required for such systems. Or consider the task of generating a summary of
a large collection of documents. With the soft attention mechanisms
used in neural systems, there is currently no direct way to condition on such an amount of text.

\section{Background}

\noindent \textbf{Summary of Notation}

\begin{framed}
\vspace{0.5em}
\hspace{-2em}
\begin{tabularx}{\linewidth}{ c c X }
  \vspace{0.2em}
 \thead{Symbol} & \thead{Shape} & \thead{Description}\\
  $V$ & scalar & size of output vocabulary\\
  $X$ & $S$ & input/source sequence $(x_1,\dots,x_{S})$ of length $S$\\
  $Y$ & $T$ & output/target sequence $(y_1,\dots,y_{T})$ of length $T$\\
  $E$ & $(S, h)$ & encoder hidden states where $E_j$ denotes representation at timestep $j$\\
  $D$ & $(T, h)$ & decoder hidden states where $D_i$ denotes representation at timestep $i$\\
  $A$ & $(T, S)$ & attention matrix where $A_{ij}$ is attention weight at $i$th decoder timestep on $j$th encoder state representation\\
  $H$ & \emph{varies} & hypothesis in hypothesis set $\mathcal{H}$ during decoding\\
  $s(H)$ & scalar & score of hypothesis $H$ during beam search\\
\end{tabularx}
\end{framed}

\noindent Minibatch size dimension is omitted from the shape column.

\subsection{Setting}

We consider modeling discrete sequences of text tokens.
Given a sequence $U = (u_1, u_2, \dots, u_{S})$ over the vocabulary $V$,
we seek to model
\begin{equation}
p(U) = \prod_{t=1}^{S} p(u_t|u_{<t})
\label{eq:lm}
\end{equation}
where $u_{<t}$ denotes $u_1, u_2, \dots, u_{t-1}$, and equality follows from the chain rule of probability.
Depending on how we choose to tokenize the text, the vocabulary can contain the
set of characters, word-pieces/byte-pairs, words, or some other unit.
For the tasks we consider in this paper, we divide the sequence $U$ into an input or \emph{source}
sequence $X$ (that is always provided in full)
and an output or \emph{target} sequence $Y$. For example, for machine translation
tasks $X$ might be a sentence in English and $Y$ the translated sentence in Chinese.
In this case, we model
\begin{equation}
p(Y|X) = \prod_{t=1}^{T} p(y_t|X, y_{<t})
\end{equation}

\begin{table}[t]
  \begin{tabularx}{\linewidth}{ l l l}
   \toprule
   \thead{Task} & \thead{$X$} (example) & \thead{$Y$} (example)\\
   \midrule
   language modeling & none (empty sequence) & tokens from news corpus\\
   machine translation & source sequence in English & target sequence in French\\
   grammar correction & noisy, ungrammatical sentence & corrected sentence\\
   summarization & body of news article & headline of article\\
   dialogue & conversation history & next response in turn\\
   \midrule
   \multicolumn{3}{c}{\textit{Related tasks (may be outside scope of this guide)}}\\
   \midrule
   speech transcription & audio / speech features & text transcript\\
   image captioning & image & caption describing image\\
   question answering & supporting text + knowledge base + question & answer\\
   \bottomrule
  \end{tabularx}
  \caption{Example tasks we consider.}
  \label{tab:tasks}
\end{table}

Note that this is a generalization of (\ref{eq:lm}); we consider $p(Y|X)$
from here on. Besides machine translation, this also encompasses many
other tasks in natural language processing---see Table~\ref{tab:tasks} for a summary.

Beyond the tasks described in the first half of Table~\ref{tab:tasks},
many of the techniques described in this paper also extend to tasks at
the intersection of text and other modalities. For example, in speech recognition,
$X$ may be a sequence of features computed on short snippets of audio,
with $Y$ being the corresponding text transcript, and in image captioning
$X$ is an image (which is not so straightforward to express as a sequence)
and $Y$ the corresponding text description. While we
could also include \emph{sequence labeling} (for example part-of-speech tagging)
as another task, we instead consider tasks that do not have a clear one-to-one
correspondence or between source and target. The lack of such a correspondence
leads to issues in decoding which we focus on in Section~\ref{sec:decoding}.
The same reasoning applies to \emph{sequence classification} tasks such as
sentiment analysis.

\subsection{Encoder-decoder models}

Encoder-decoder models, also referred to as sequence-to-sequence models,
were developed for machine translation and have rapidly exceeded the performance
of prior systems depite having comparatively simple architectures,
trained end-to-end to map source directly to target.

\begin{figure}[t]
  \centering
  \includegraphics[]{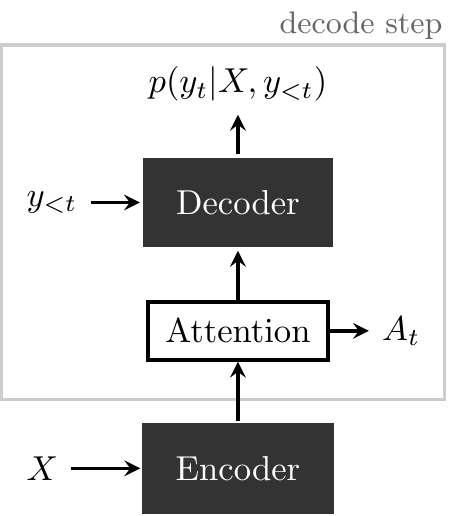}
  \caption{Figure illustrating the generic encoder-decoder model architecture
    we assume for this guide. Several choices are possible for the
    encoder and decoder architectures as well as for the attention mechanism.
    Here we show the outputs for a single timestep.}
  \label{fig:encdec}
\end{figure}

Before neural network-based approaches, count-based methods \citep{chen1996empirical} and
methods involving learning phrase pair probabilities were used for
language modeling and translation.
Prior to more recent encoder-decoder models,
\emph{feed-forward fully-connected neural networks}
were shown to work well for language modeling. Such models simply stack affine
matrix transforms followed by nonlinearities to the input and each following hidden
layer~\citep{bengio2003neural}. However, these networks have fallen out of
favor for modeling sequence data, as they require defining a fixed context
length when modeling $p(y_t|X,y_{<t})$, do not use parameter sharing across timesteps,
and have been surpassed in performance by subsequent architectures.

At the time of this writing, several different architectures have demonstrated
strong results.
\begin{itemize}
  \item \emph{Recurrent neural networks} (RNNs) use shared parameter matrices across
different time steps and combine the input at the current time step with the
previous hidden state summarizing all previous time steps~\citep{mikolov2010,sutskever2014sequence,cho2014learning}.
Many different gating mechanisms have been developed for such architectures
to try and ease optimization~\citep{hochreiter1997long,cho2014learning}.
\item \emph{Convolutional neural networks} (CNNs). Convolutions with kernels reused across timesteps
can also be used with masking to avoid peeking ahead at
future inputs during training (see Section~\ref{sec:train_overview} for an overview
of the training procedure) \citep{kalchbrenner2016neural}.
Using convolutions has the benefit during training
of parallelizing across the time dimension instead of computing the next hidden state
one step at a time.
\item Both recurrent and convolutional networks for modeling sequences typically
  rely on a per timestep \emph{attention} mechanism \citep{bahdanau2014neural} that acts as a shortcut connection between
the target output prediction and the relevant source input hidden states.
At a high-level, at decoder timestep $i$, the decoder representation $D_i$ is
used to compute a weight $\alpha_{ij}$ for each encoder representation $E_j$.
For example, this could be done by using the dot product $D_i^\top E_j$
as the logits before applying the softmax function. Hence
$$\alpha_{ij} =  \exp(D_i^\top E_j) / \sum_{k=1}^S \exp(D_i^\top E_k)$$
The weighted representation $\sum_{j=1}^S \alpha_{ij} E_j$ is then fed into
the decoder along with $X$ and  $y_{<i}$.
More recent models which rely purely on attention mechanisms with masking
have also been shown to obtain as good or better results as RNN or CNN-based models~\citep{vaswani2017attention}.
We describe the attention mechanism in more detail in Section~\ref{sec:attention}.
\end{itemize}

\noindent Unless otherwise indicated, the advice in this guide is intended to be
agnostic of the model architecture, as long as the following two conditions hold:
\begin{enumerate}
  \item The model performs next-step prediction of the next target conditioned on the source and previous predicted targets, i.e. it models $p(y_t|X,y_{<t})$.
  \item The model uses an attention mechanism (resulting in an attention matrix $A$), which eases training, is simple to implement and cheap to compute in most cases, and has become a standard component of encoder-decoder models.\footnote{Some recent architectures also make use of a \emph{self-attention} mechanism where decoder outputs are conditioned on previous decoder hidden states; for simplicity we do not discuss this extension.}
\end{enumerate}
Figure~\ref{fig:encdec} illustrates the backbone architecture we use for this guide.

\subsection{Training overview}
\label{sec:train_overview}

During training, we optimize over the model parameters $\theta$ the
sequence cross-entropy loss
\begin{equation}
  \ell(\theta) = -\sum_{t=1}^T \log p(y_t|X, y_{<t}; \theta).
\end{equation}
thus maximizing the log-likelihood of the training data.
Previous ground truth inputs are given to the model when predicting the
next index in the sequence, a training method sometimes referred to (unfortunately) as
\emph{teacher forcing}.
Due to the inability to fit current datasets into memory
as well as for faster convergence,
gradient updates are computed on minibatches of training sentences.
\emph{Stochastic gradient descent} (SGD) as well as optimizers such as Adam \citep{kingma2014adam} have been shown to
work well empirically.

Recent reserarch has also explored other methods for training sequence models, such
as by using reinforcement learning or a separate adversarial
loss~\citep{goodfellow2014generative,li2016deep,li2017adversarial,bahdanau2016actor,arjovsky2017wasserstein}.
As of this writing, however, the aforementioned training method
is the primary workhorse for training such models.

\subsection{Decoding overview}
\label{sec:dec_overview}

During decoding, we are given the source sequence $X$ and seek to generate the
target $\hat{Y}$ that maximizes some scoring function $s(\hat{Y})$.\footnote{The scoring
  function may also take $X$ or other tensors as input, but for simplicity we consider just $\hat{Y}$.}
In \emph{greedy decoding}, we simply take the argmax over the softmax output distribution for each timestep, then feed that as the input for the next timestep.
Thus at any timestep we only have a single hypothesis.
Although greedy decoding can work surprisingly well, note that it often does \emph{not} result in the most probable output hypothesis, since there may be a path that is more probable overall despite including an output which was not the argmax (this also holds true for most scoring functions we may choose).

Since it's usually intractable to consider every possible $\hat{Y}$ due to
the branching factor and number of timesteps,
we instead perform \emph{beam search}, where we iteratively expand each hypotheses one
token at a time, and at the end of every search iteration we only keep the $k$
best (in terms of $s(\cdot)$) hypotheses, where $k$ is the \emph{beam width} or \emph{beam size}.
Here's the full beam search procedure, in more detail:

\begin{enumerate}
  \item We begin the beam procedure with the start-of-sequence token, \sos.
    Thus our set of hypothesis $\mathcal{H} = \{[\mathrmsos]\}$ consisting of
    the single hypothesis $H = [\mathrmsos]$, a list with only the start token.
  \item Repeat, for $t=1,2,\dots,T_{\mathrm{max}}$:
  \begin{enumerate}
    \item Repeat, for $H \in \mathcal{H}$:
      \begin{enumerate}
        \item Repeat, for every $u$ in the vocabulary $V$ with probability $p_{ut} = p(u|X,H)$:
          \begin{enumerate}
            \item Add the hypothesis $H_\mathrm{new} = [\mathrmsos, h_1, \dots, h_{t-1}, u]$ to $\mathcal{H}$.
            \item Compute and cache $s(H_\mathrm{new})$. For example, if $s(\cdot)$ simply computes the cumulative log-probability of a hypothesis, we have $s(H_\mathrm{new}) = s(H) + \log p_{ut}$.
          \end{enumerate}
        \end{enumerate}
    \item If any of the hypotheses end with the end-of-sequence token \eos, move
      that hypothesis to the list of terminated hypotheses $\mathcal{H}_\mathrm{final}$.
    \item Keep the $k$ best remaining hypotheses in $\mathcal{H}$ according to $s(\cdot)$.\footnote{Likewise, we can also store other information for each hypothesis, such as $A_t$.}
  \end{enumerate}
\item Finally, we return $H^{*} = \mathrm{arg}\max_{H\in\mathcal{H}_\mathrm{final}} s(H)$.
      Since we are now considering completed hypotheses, we may also wish to use
      a modified scoring function $s_\mathrm{final}(\cdot)$.
\end{enumerate}

\begin{figure}[t]
  \centering
  \includegraphics{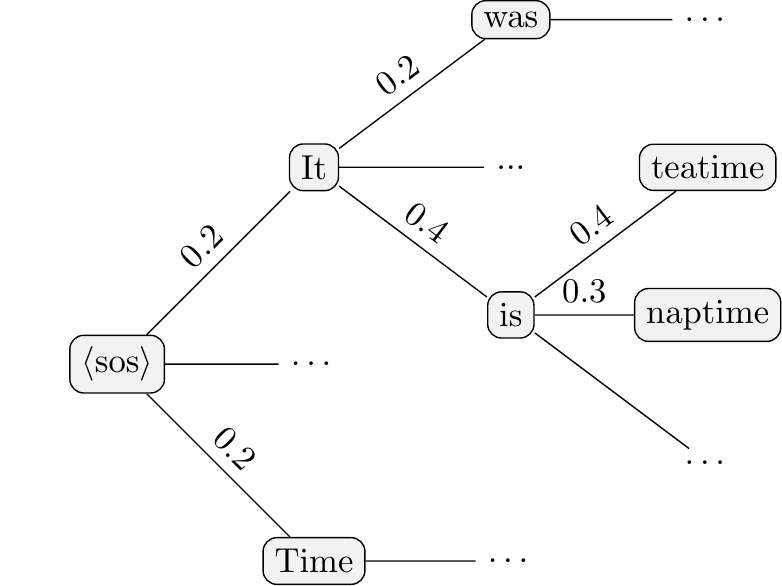}
  \caption{Toy example of beam search procedure with beam width $k=2$.
    The search has been run for 3 steps and no hypothesis has terminated with \eos yet.
    Edges are annotated with probabilities of tokens.
    Only the tokens after pruning to the top $k$ hypotheses are shown.}
\end{figure}

\noindent One surprising result with neural models is that relatively small beam sizes yield good results with rapidly diminishing returns. Further, larger beam sizes can even yield (slightly) worse results. For example, a beam size of 8 may only work marginally better than a beam size of 4, and a beam size of 16 may work worse than 8~\citep{koehn2017six}.

Finally, oftentimes incorporating a language model (LM) in the scoring function
can help improve performance.
Since LMs only need to be trained on the target corpus, we can train language
models on much larger corpuses that do not have parallel data.
Our objective in decoding is to maximize the joint probability:
\begin{align}
  \mathrm{arg}\max_{\hat{Y}} p(X, \hat{Y}) &= \mathrm{arg}\max_{\hat{Y}} p(X|\hat{Y}) p(\hat{Y})
\end{align}
However, since we are given $X$ and not $Y$, it's intractable to maximize
over $p(X|\hat{Y})$. In practice, we instead maximize the pseudo-objective
$$
  \mathrm{arg}\max_{\hat{Y}} p(\hat{Y}|X) p(\hat{Y}).
$$

If our original score is $s(H)$, which we assume involves a $p(H|X)$ term, then the LM-augmented score is
$$s(H) + \lambda \log p_\mathrm{LM}(H)$$
where $\lambda$ is a hyperparameter to balance the LM and decoder scores.


Despite these issues, this simple procedure works fairly well. However, there arise many cases where beam search, with or without a language model, can result in far from optimal outputs. This is due to inherent biases in the decoding (as well as the training) procedure. We describe how to diagnose and tackle these problems that can arise in Section~\ref{sec:decoding}.

\subsection{Attention}
\label{sec:attention}

\begin{figure}[h]
  \centering
  \includegraphics[scale=0.7]{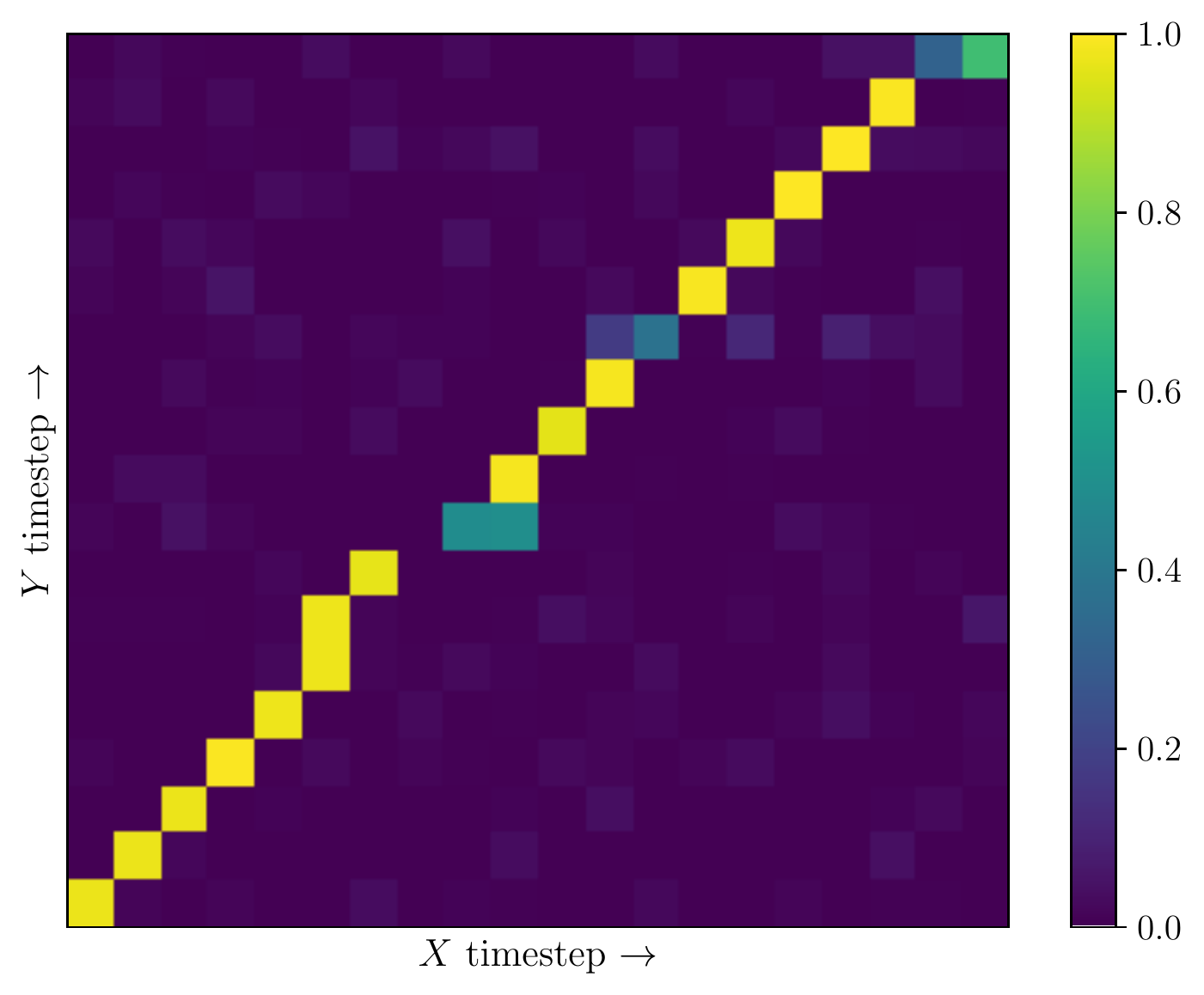}
  \caption{Expected attention matrix $A$ when source and target are monotonically aligned (synthesized as illustrative example).}
  \label{fig:good_att_mat}
\end{figure}

The basic attention mechanism used to ``attend'' to portions of the encoder hidden
states during each decoder timestep has many extensions and applications.
Attention can also be over previous decoder hidden states, in what is called
self-attention~\citep{vaswani2017attention}, or used over components separate from the encoder-decoder model
instead of over the encoder hidden states~\cite{grave2016improving}. It can also be used to monitor
training progress (by inspecting whether a clear alignment develops between encoder
and decoder hidden states), as well as to inspect the correspondences between
the input and output sequence that the network learns.

That last bit---that the attention matrix $A$ typically follows the correspondences
between input and output---will be useful when we discuss methods for guiding
the decoding procedure in Section~\ref{sec:decoding}. Hence we go into more detail
on the attention matrix $A$ here.

The attention matrix  $A$ has $T$ columns and $S$ rows, where $T$ is the number of
output timesteps and $S$ is the number of input timesteps.
Every row $A_i$ is a discrete probability distribution over the encoder hidden states,
which in this guide we assume to be equal to the number of input timesteps $S$.
In the case that a encoder column $A_{:j}$ has no entry
over some threshold (say $0.5$), this suggests that the corresponding
encoder input was ignored by the decoder. Likewise, if $A_{:j}$ has
multiple values over threshold, this suggests those encoder hidden states were
repeatedly used during multiple decoder timesteps.
Figure~\ref{fig:good_att_mat} shows an attention matrix we might expect
for a well-trained network where source and target are well-aligned
(e.g. English$\rightarrow$French translation).
For a great overview and visualizations of of attention and RNN models,
also see~\cite{olah2016attention}.

\subsection{Evaluation}

One of the key challenges for developing text generation systems is that
there is no satisfying automated metric for evaluating the final output of the system.
Unlike classification and sequence labeling tasks, we (as of now) cannot
precisely measure the output quality of text generation systems
barring human evaluation. Perplexity does not always correlate well with
downstream metrics~\citep{chen1998evaluation}, automated or otherwise.
Common metrics based on $n$-gram overlap such as ROUGE~\citep{lin2004rouge} and BLEU~\citep{papineni2002bleu} are only rough approximations, and often do not capture linguistic fluency and cohererence~\citep{conroy2008mind,liu2016not}.
Such metrics are especially problematic in more open-ended generation tasks such as summarization
and dialogue.

Recent results have shown that though automated metrics are not great at
distinguishing between systems once performance passes some baseline,
they nonetheless are useful for finding examples where performance is poor,
and can also be consistent for evaluating similar systems \citep{novikova2017we}.
Thus, despite issues with current automated evaluation metrics, we assume their use for model development;
however,  manual human evaluation should be interspersed as well.

\section{Preprocessing}
\label{sec:preproc}

With increasingly advanced libraries for building computation graphs
and performing automatic differentiation, a more significant portion
of the software development process is devoted to data
preparation.\footnote{Suggestions for multilingual preprocessing are welcome.}

Broadly speaking, once the raw data has been collected, there remains
cleaning, tokenization, and splitting into training and test data.
An important consideration during cleaning is setting the
character encoding---for example ASCII or UTF-8---for
which libraries such as Python's \texttt{unidecode} can save a lot of time.
After cleaning comes the less easily-specified tasks of splitting the
text into sentences and tokenization. At present, we recommend
Stanford CoreNLP\footnote{\url{https://github.com/stanfordnlp/CoreNLP}}
for extensive options and better handling of sentence and word boundaries
than other available libraries.

An alternative to performing tokenization (and later detokenization)
is to avoid it altogether. Instead of working at the word level,
we can instead operate at the character level or use intermediate
subword units \citep{sennrich2015neural}. Such models result in
longer sequences overall, but empirically subword models tend to
provide a good trade-off between sequence length (speed) and handling
of rare words~\citep{wu2016google}. Section~\ref{ssec:oovs} discusses
the benefits of subword models in more detail.
Ultimately, if using word tokens, it's important to use a consistent
tokenization scheme for all inputs to the system---this includes
handling of contractions, punctuation marks such as quotes and hyphens,
periods denoting abbreviations (nonbreaking prefixes) vs. sentence boundaries, character
escaping, etc.\footnote{The Stanford Tokenizer page \url{https://nlp.stanford.edu/software/tokenizer.html} has a detailed list of options.}

\section{Training}

A few heuristics should be sufficient for handling many of the issues when
training such models. Start by getting the model to overfit on a tiny subset
of the data as a quick sanity check. If the loss explodes, keep reducing the learning rate
until it doesn't. If the model overfits, apply dropout~\citep{srivastava2014dropout,zaremba2014recurrent} and weight
decay until it doesn't.
Gradient clipping is often crucial to avoid the exploding gradient
  problem while using a reasonably large learning rate.
For SGD and its variants, periodically annealing the learning rate
when the validation loss fails to decrease typically helps significantly.

A few useful heuristics that should be robust to the
hyperparameter settings and optimization settings you use:
\begin{itemize}
  \item Sort the next dozen or so batches of sentences by length so each batch has
    examples of roughly the same length, thus saving computation~\citep{sutskever2014sequence}.
  \item If the training set is small, tuning regularization will be key to
    performance~\citep{melis2017state}. Noising (or ``token dropout") is also worth trying \citep{xie2017data}.
  Though we only touch on this issue briefly, amount of training data will
  be in most cases the primary bottleneck in the performance of a NTG model.
  \item Measure validation loss after each epoch and anneal the learning rate
  when validation loss stops decreasing. Depending on how
  much the validation loss fluctates (based off of validation set size and optimizer
  settings) you may wish to anneal with patience (wait for several epochs of non-decreasing learning rate before reducing the learning rate).
  \item Periodically checkpoint model parameters and measure downstream
  performance (BLEU, $F_1$, etc.) using several of the last few model checkpoints.
  Validation cross entropy loss and final performance may not correlate well, and there can
  be significant differences in final performance across checkpoints with similar
  validation losses.
  \item Ensembling almost always improves performance. Averaging checkpoints
    is a cheap way to approximate the ensembling effect~\citep{huang2017snapshot}.
\end{itemize}

\noindent For a survey of model parameters to consider as well as suggested settings of hyperparameters,
see \cite{britz2017massive} or \cite{melis2017state}.

\section{Decoding}
\label{sec:decoding}

Suppose you've trained a neural network encoder-decoder model that achieves
reasonable perplexity on the validation set. You then try running decoding
or generation using this model. The simplest way
is to run greedy decoding, as described in Section~\ref{sec:dec_overview}.
From there, beam search decoding should yield some additional performance
improvements. However, it's rare that things simply work.
This section is intended for use as a quick reference when encountering
common issues during decoding.\footnote{Many examples are purely illustrative excerpts from \emph{Alice's Adventures in Wonderland}~\citep{carroll1865alice}.}

\subsection{Diagnostics}

First, besides manual inspection, it's helpful to create some
diagnostic metrics when debugging the different components
of a text generation system.
Despite training the encoder-decoder network to map source to target,
during the decoding procedure we introduce two additional components:
\begin{enumerate}
  \item A scoring function $s(H)$ that tells us how ``good" a hypothesis $H$ on the beam is.
  \item Optionally, a language model trained on a large corpus which may or may not be similar to the target corpus.
\end{enumerate}

It may not be clear which of these components we should prioritize
when trying to improve the performance of the combined system;
hence it can be very helpful to run ablative analysis \citep{ngadvice},

For the language model, a few suggestions are measuring
performance for $\lambda=0$ and several other reasonably spaced values,
then plotting the performance trend; measuring perplexity of the language
model when trained on varying amounts of training data, to see
if more data would be helpful or yields diminishing returns; and
measuring performance when training the language model on several different
domains (news data, Wikipedia, etc.)
in cases where it's difficult to obtain data close to the target domain.

When measuring the scoring function, computing metrics and inspecting
the decoded outputs vs. the gold sentences often immediately yields insights.
Useful metrics include:
\begin{itemize}
  \item Average length of decoded outputs $\hat{Y}$ vs. average length of reference targets $Y$.
  \item $s(\hat{Y})$ vs. $s(Y)$, then inspecting the ratio $s(\hat{Y})/s(Y)$.
    If the average ratio is especially low, then there may be a bug in the beam search,
    or the beam size may need to be increased.  If the average ratio is high,
    then the scoring function may not be appropriate.
  \item For some applications computing edit distance (insertions, substitutions, deletions) between $\hat{Y}$ and $Y$ may also be useful, for example by looking at the most frequent edits or by examining cases where the length-normalized distances are highest.
\end{itemize}

\subsection{Common issues}

\subsubsection{Rare and out-of-vocabulary (OOV) words}
\label{ssec:oovs}
\begin{samepage}
\begin{framed}
  \noindent
  \begin{tabular}{r l}
    \textbf{Decoded} & And as in \unk\ thought he stood, The \unk, with eyes of flame\eos\\
    \textbf{Expected} & And as in uffish thought he stood, The Jabberwock, with eyes of flame\eos
  \end{tabular}
  \label{tab:oov-ex}
\end{framed}
\end{samepage}

For languages with very large vocabularies, especially languages with rich
morphologies, rare words become problematic when choosing a tokenization scheme
that results in more token labels than it is feasible to model in the
output softmax. One ad hoc approach that was first used to
deal with this issue is simply
to truncate the softmax output size (to say, 50K), then assign the
remaining token labels all to the \unk\ class~\citep{luong2014addressing}.
The box above illustrates the resulting output (after detokenization) when rare words are replaced with \unk s.
A more elegant approach is to use
character or subword preprocessing~\citep{sennrich2015neural,wu2016google}
to avoid OOVs entirely, though this can slow down runtime for both training and decoding.

\subsubsection{Decoded output short, truncated or ignore portions of input}

\begin{samepage}
\begin{framed}
  \noindent
  \begin{tabular}{r l}
    \textbf{Decoded} & It's no use going back to yesterday.\eos \\
    \textbf{Expected} & It's no use going back to yesterday, because I was a different person then.\eos
  \end{tabular}
\end{framed}
\end{samepage}

During the decoding search procedure, hypotheses terminate with the \eos token.
The decoder network should learn to place very low probability on the \eos
token until the target is fully generated; however, sometimes \eos does not
have sufficiently low probability. This is because as the length of the
hypothesis grows, the total log probability only decreases. Thus, if we
do not normalize the log probability by the length of the hypothesis, shorter
hypotheses will be favored. The box above illustrates an example where the
hypothesis terminates early.
This issue is exacerbated when incorporating a
language model term. Two simple ways of resolving this issue are normalizing
the log-probability score and adding a length bonus.
\begin{itemize}
  \item \emph{Length normalization}: Replace the score $s(\hat{Y})$ with
    the score normalized by the hypothesis length $s(\hat{Y})/\hat{T}$.
  \item \emph{Length bonus}:  Replace the score $s(\hat{Y})$ with
    the $s(\hat{Y}) + \beta\hat{T}$, where $\beta > 0$ is a hyperparameter.
\end{itemize}
Note that normalizing the total log-probability by length is equivalent to
maximizing the $\hat{T}$\textsuperscript{th} root of the probability, while
adding a length bonus is equivalent to multiplying the probability at every
timestep by a baseline $e^\beta$.

\begin{figure}
  \centering
  \includegraphics[scale=0.7]{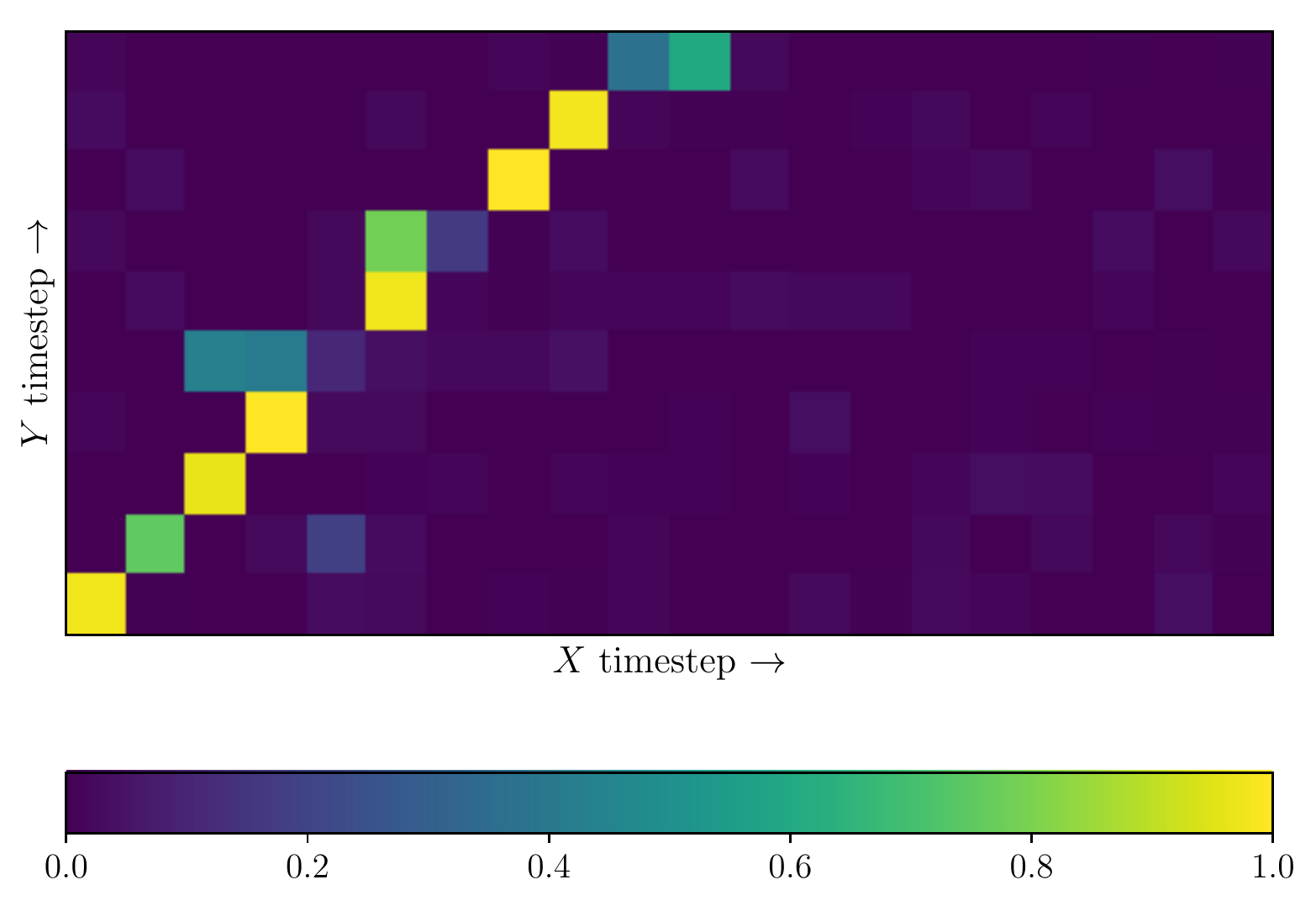}
  \caption{Example of attention matrix $A$ when decoding terminates early with the \eos token without having covered the input $X$.}
\end{figure}

Another method for avoiding this issue is with a \emph{coverage penalty}
using the attention matrix $A$~\citep{tu2016modeling,wu2016google}.
As formulated here, the coverage penalty can only be applied once a hypothesis
(with a corresponding attention matrix) has terminated; hence it can
only be incorporated into $s_\mathrm{final}(\cdot)$ to perform a final re-ranking
of the hypotheses. For a given hypothesis $H$ with attention matrix $A$ with
shape $(\hat{T}, S)$, the coverage penalty is computed as
\begin{equation}
  \mathrm{cp}(A) = \sum_{j=1}^S \log \left[ \min \left( \sum_{i=1}^{\hat{T}} A_{ij}, 1.0 \right) \right]
\end{equation}
Intuitively, for every source timestep, if the attention matrix places full
probability on that source timestep when aggregated over all decoding timesteps, then the coverage penalty is zero. Otherwise a penalty is incurred for not attending to that source timestep.

Finally, when the source and target are expected to be of roughly equal lengths,
one last trick is to simply constrain the target
length $\hat{T}$ to be within some delta of the source length $S$,
e.g. $\max(S-\delta, (1-\Delta)\cdot S) \le \hat{T} \le \max(S+\delta, (1+\Delta)\cdot S)$ where we also
  include $S-\delta$ and $S+\delta$ in the case where $S$ is very small, and $\delta$
  and $\Delta$ are hyperparameters.

\subsubsection{Decoded output repeats}

\begin{samepage}
\begin{framed}
  \noindent
  \begin{tabular}{r l}
    \textbf{Decoded} & I'm not myself, you see, you see, you see, you see, \dots \\
    \textbf{Expected} & I'm not myself, you see.\eos
  \end{tabular}
\end{framed}
\end{samepage}

\begin{figure}
  \centering
  \includegraphics[scale=0.7]{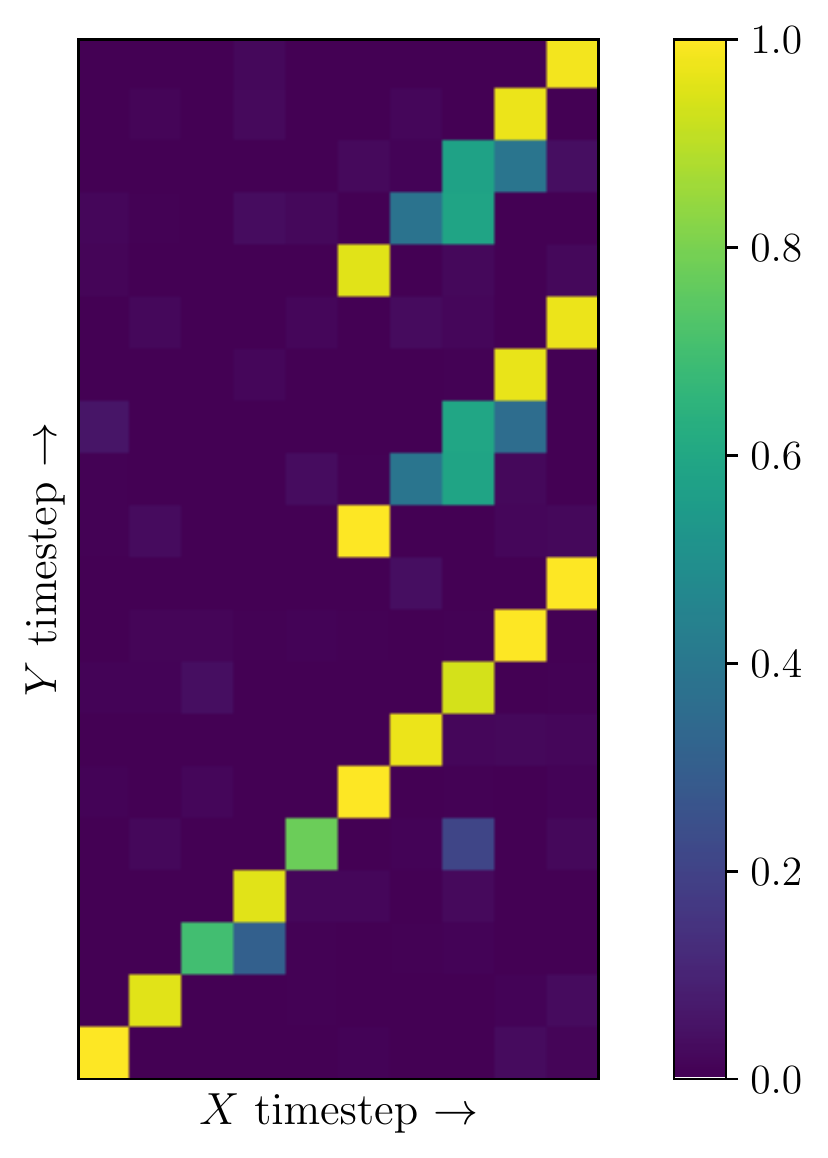}
  \caption{Example of attention matrix $A$ when decoding exhibits repeating behavior.}
\end{figure}

Repeating outputs are a common issue that often seem to expose
neural versus template-based systems. Simple measures include
adding a penalty when the model reattends to previous timesteps
after the attention has shifted away. This is easily detected using
the attention matrix $A$ with some manually selected threshold.

Finally, a more fundamental issue to consider with repeating outputs
is poor training of the model parameters. Passing in the attention
vector $A_{i-1}$ as part of the decoder input when predicting $y_i$ is
another training-time method~\citep{see2017get}.

\subsubsection{Lack of diversity}

\begin{framed}
  \noindent ``I don't know"~\citep{li2015diversity}
\end{framed}

\noindent
In dialogue and QA, where there are often very common responses for many
  different conversation turns, generic responses such as ``I don't know"
  are a common problem.
Similarly, in problems where many possible source inputs map to a much
smaller set of possible target outputs, diversity of outputs
can be an issue.

Increasing the temperature $\tau$ of the softmax
$\exp(z_i/\tau) / \sum_j \exp(z_j/\tau)$ is a simple
method for trying to encourage more diversity in decoded
outputs. In practice, however, a method penalizing
low-ranked siblings during each step of the beam search decoding
procedure has been shown to work well~\citep{li2016simple}.
Another more sophisticated method is to maximize the mutual information
between the source and target, but is significantly more difficult
to implement and requires generating $n$-best lists~\citep{li2015diversity}.


\subsection{Deployment}
\label{ssec:deploy}

Although speed of decoding is not a huge concern when trying to achieve
state-of-the-art results, it is a concern when deploying models in production,
when real-time decoding is often a requirement.
Beyond gains from using highly parallelized hardware such as GPUs or
from using libraries with optimized matrix-vector operations,
we now discuss some other techniques for improving the runtime of decoding.

Consider the factors which determine the runtime of the decoding algorithm.
For the beam search algorithms we consider, the runtime should scale linearly
with the beam size $k$ (although in practice, batching hypotheses can lead to
sublinear scaling). The runtime will often scale approximately quadratically with
the hidden size of network $n$, and finally, linearly with the number of timesteps $t$.
Thus decoding might have a complexity of $O(kn^2t)$.

Thus, (a jumbled collection of)
possible methods for speeding up decoding include developing
heuristics to prune the beam, finding the best trade-off between size
of the vocabulary (softmax) and decoder timesteps, batching multiple
examples together, caching previous computations (in the case of CNN models),
and performing as much computation as possible within the compiled computation graph.

\section{Conclusion}

We describe techniques for training and dealing with undesired behavior in natural language generation models using neural network decoders. Since training models tends to be far more time-consuming than decoding, it is worth making sure your decoder is fully debugged before committing to training additional models. Encoder-decoder models are evolving rapidly, but we hope these techniques will be useful for diagnosing a variety of issues when developing your NTG system.

\pagebreak

\section{Acknowledgements}

 We thank Arun Chaganty and Yingtao Tian for helpful discussions,
 and Dan Jurafsky for many helpful pointers.

\setstretch{1.0}
\bibliographystyle{abbrvnat}
\bibliography{refs.bib}

\end{document}